\documentclass{article}
\usepackage[square,numbers]{natbib}
\usepackage[symbol]{footmisc}

\usepackage{authblk}
\usepackage[top=10mm, bottom=20mm, left=20mm, right=20mm]{geometry}
\usepackage{multirow}
\usepackage{bbold}
\usepackage{multicol}
\usepackage{lipsum}

\usepackage{graphicx}
\usepackage{caption}
\usepackage{subcaption}
\usepackage{hyperref}
\hypersetup{hidelinks,
backref=true,
pagebackref=true,
hyperindex=true,
breaklinks=true,
colorlinks=true,
linkcolor=blue,
urlcolor=blue,
pdftitle={Title},
pdfauthor={Author}}

\usepackage{float}

\begin{document}

\title{\textbf{Adversarial Attacks on Deep Models for Financial Transaction Records}}

\author[1]{Ivan Fursov\footnote{Equal contribution}}

\newcommand\CoAuthorMark{\footnotemark[\arabic{footnote}]} 
\author[1]{Matvey Morozov\protect\CoAuthorMark
}
\affil[1]{%
  Skoltech, Moscow, Russia
}
\author[1,2,4]{Nina Kaploukhaya}
\affil[2]{
  IITP RAS, Moscow, Russia
%
}

\author[1]{Elizaveta Kovtun}

\author[1]{Rodrigo Rivera-Castro}

\author[3,4]{Gleb Gusev  \thanks{E-mail: gleb57@gmail.com}}
\author[3]{Dmitry Babaev}
\author[3]{Ivan Kireev}
\affil[3]{%
  Sber AI lab, Moscow, Russia
  }
\affil[4]{%
  MIPT, Moscow, Russia
  }

\author[1]{Alexey Zaytsev${}^*$
\thanks{E-mail: a.zaytsev@skoltech.ru}
}
\author[1]{Evgeny Burnaev}
\date{\vspace{-4ex}}

\maketitle

\newcommand{\GG}[1]{\textcolor{red}{[GG: #1]}}

\newcommand{\MM}[1]{\textcolor{red}{[Matvey: #1]}}


\setlength{\columnsep}{0.5cm} 
\begin{multicols}{2} 

\section*{Abstract}
 \vspace{-3mm} 
Machine learning models using transaction records as inputs are popular among financial institutions. 
The most efficient models use deep-learning architectures similar to those in the NLP community, posing a challenge due to their tremendous number of parameters and limited robustness. 
In particular, deep-learning models are vulnerable to adversarial attacks: a little change in the input harms the model's output. 
  
In this work, we examine adversarial attacks on transaction records data and defences from these attacks. 
The transaction records data have a different structure than the canonical NLP or time series data, as neighbouring records are less connected than words in sentences, and each record consists of both discrete merchant code and continuous transaction amount.
We consider a black-box attack scenario, where the attack doesn't know the true decision model, and pay special attention to adding transaction tokens to the end of a sequence.
These limitations provide more realistic scenario, previously unexplored in NLP world.

The proposed adversarial attacks and the respective defences demonstrate remarkable performance using relevant datasets from the financial industry. 
Our results show that a couple of generated transactions are sufficient to fool a deep-learning model. 
Further, we improve model robustness via adversarial training or separate adversarial examples detection.
This work shows that embedding protection from adversarial attacks improves model robustness, allowing a wider adoption of deep models for transaction records in banking and finance.



 \vspace{-5mm} 
\section{Introduction}
 \vspace{-3mm} 

Adversarial attacks are a fundamental problem in deep learning. 
Alone a small targeted perturbation in the machine learning model's input results in a wrong prediction by the model~\cite{goodfellow2014explaining}.
These type of attacks are pervasive and present in various domains such as computer vision, natural language processing, event sequence processing, and graphs~\cite{chakraborty2018adversarial}. 

\begin{figure}[H]
    \centering
    \includegraphics[width=0.5\textwidth]{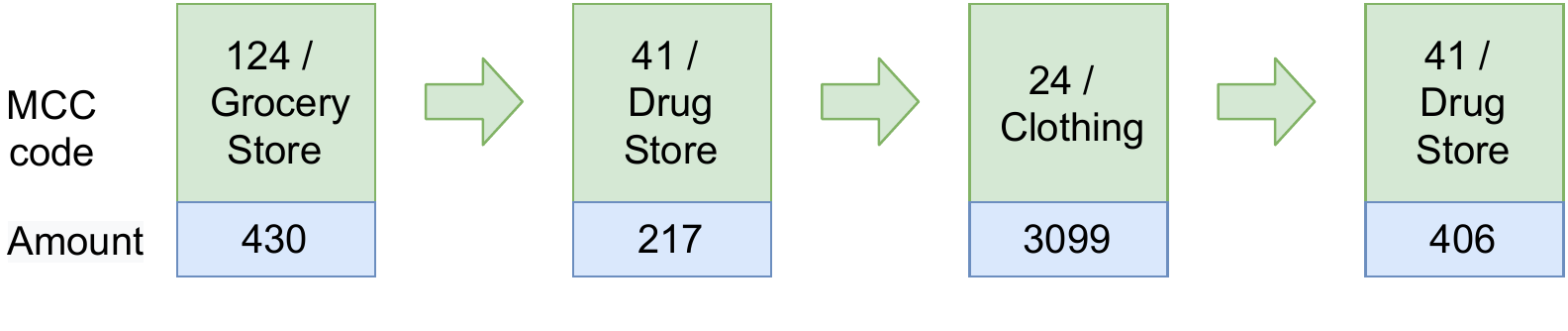}
    \caption{Example of a transaction records data sequence: Each transaction has information about its MCC (Merchant Category Codes) describing the purchase category and purchase amount. Data can also include transaction location and time values.}
    \label{fig:trans_ex}
\end{figure}

In this study, we consider a specific application domain of data models based on transaction records similar to~\autoref{fig:trans_ex}.
This type of data arises in the financial industry, a sector that sees a vertiginous adoption of deep-learning models with a large number of parameters ~\cite{babaev2019rnn,babaev2020metric}.
Analysts use them to detect a credit default or fraud.
As a result, deep-learning models on sequences of transactions stemming from a client or a group of clients are a natural combination.
However, given their vast number of parameters, these models can be vulnerable to adversarial attacks~\cite{goodfellow2014explaining}.
Hence, risk mitigation related to the possibility of adversarial attacks becomes paramount.
Despite the great attention of both academy and industry to adversarial attacks in machine learning, there are still no papers studying this problem in the finance domain.
We must make these models robust to such attacks if we want to see them widely adopted by the industry.

Transaction records data consists of sequences, making it possible to transfer some techniques from the domain of natural language processing~\cite{zhang2019adversarial}. 
However, there are several peculiarities related to the generation of adversarial sequences of transactions:
\begin{enumerate}
    \item Unlike neighbouring words in natural language, there may be no logical connection between neighbouring transactions in a sequence.
\item An attacker cannot insert a token at an arbitrary place in a sequence of transactions in most cases. Often, the attacker can only add new transactions to the end of a sequence.
\item The concept of semantic similarity is underdeveloped when we compare it to what we see in the NLP world~\cite{zhang2019bertscore}. 
\item Transaction data is complex. Besides MCC, which is categorical, we also have the transaction amount, and can also have its location, timestamp, and other fields. It is particularly an essential question if adversarial transactions' amount affects an attack's success.
\end{enumerate}

To make an attack scenario realistic, we consider inference-time black-box attacks, i.e. an attacker has access only to a substitute model different from the target attacked model and trained using a different dataset and attacks a model, when it is used in production. We compare this scenario to a more dangerous but less realistic white-box attack.
This setting is close to works such as~\cite{fursov2020gradient} for NLP and~\cite{szegedy2013intriguing} for CV models.
 \vspace{-3mm} 
\subsection{Novelty}
\begin{figure}[H]
    \centering
    \includegraphics[width=0.375\textwidth]{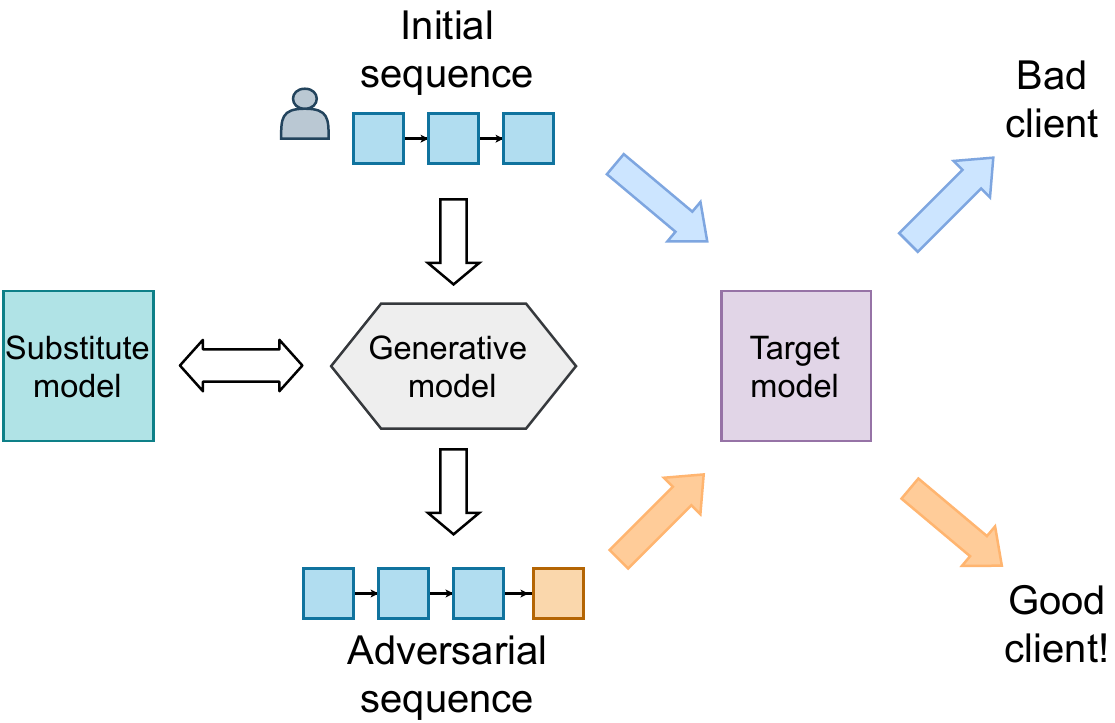}
    \caption{Proposed attack scheme: we consider a realistic black box scenario, when a malicious model user has access to a separate substitute model during an attack on a target model. To fool a target model, he adds new transactions to the end of a sequence making additional purchases. To select new sequence, the attacker adopts a conditioned generative model.}
    \label{fig:general_attack_scheme}
\end{figure}

\vspace{-2mm} 
This work addresses three research questions: 
(1) The vulnerability of finance models to different attack methods in a realistic scenario, 
(2) the effectiveness of adversarial training, 
(3) the role of the amount of adversarial transactions.
We propose several black-box adversarial attacks for transaction records data for realistic scenarios and provide defences from these attacks accounting for the above challenges.
Our black-box attacks change or \emph{add} only a couple of tokens with minimal money amounts for each token to fool the model.
We adopt the idea of loss gradients and a substitute model to create effective adversarial attacks using a separately trained generative model for transactions models.
The scheme of our attack is in Figure~\ref{fig:general_attack_scheme}.

Our defences make most of these attacks ineffective besides an attack based on sampling from a language model.
Banks and other financial actors can use our results to improve their fraud detection and scoring systems.


The main claims and structure of the paper are as follows:
\begin{itemize}
    \item We address an important and practical research questions: the vulnerability of deep finance models to adversarial attacks in the most realistic black-box scenarios, and the effectiveness of adversarial learning;
    \item We develop black-box adversarial attacks for transaction records data from the financial industry and defences from these attacks in the light of the above challenges. 
    Our approach adopts the idea of loss gradients and a substitute model to create effective adversarial attacks based on a generative model that takes into account mixed disrete-continuous structure of bank transactions data. 
    \autoref{sec:attack_methods} describes our attacks and defenses.
    \item Our black-box attacks change or add only a couple of tokens with small monetary values for each token to fool the model. 
    We show that in~\autoref{subsec:overall_quality}.
    \item Our defences render most of these attacks ineffective except for an attack based on sampling from a language model. 
    We present it in~\autoref{subsec:defenses_results}.
    \item We conduct experiments on relevant datasets from the industry and provide an investigation on the effectiveness of such attacks and how one can defend her model from such attacks.
    Banks can use the obtained results to improve their fraud detection and scoring systems. 
    We provide a discussion on our attacks and defences in~\autoref{subsec:closer_look}.
\end{itemize}
 \vspace{-5mm} 
\section{Related work}
 \vspace{-3mm} 
Adversarial attacks exist for most deep-learning models in domains as wide as, ~\cite{yuan2019adversarial}, computer vision~\cite{akhtar2018threat,khrulkov2018art}, natural language~\cite{zhang2019adversarial,wang2019survey}, and graph data~\cite{sun2018adversarial}. 
A successful attack generates adversarial examples that are close to an initial input but result in a misclassification by a deep-learning model~\cite{kurakin2016adversarial}.
\cite{xu2019review,pitropakis2019taxonomy} provides a comprehensive review.
For sequential data and especially for NLP data, there are also numerous advances in recent years. 
\cite{zhang2019adversarial,wang2019survey} discuss them.
Similarly, the work of~\cite{zeager2017adversarial} emphasizes adversarial attacks in financial applications.

Our study uses transaction records data. 
As a consequence, we generate adversarial examples for discrete sequences of tokens.
For NLP data, \cite{papernot2016crafting} is one of the first approaches to deal with this challenge. 
The authors consider white-box attacks for an RNN model. 
In our work, we focus on black-box attacks. 
We consider this scenario more realistic than the white-box one. 
Along the same lines~\cite{gao2018black} also consider black-box attacks for discrete sequences. 
They present an algorithm, DeepWordBug, to identify the most relevant tokens for the classifier whose modifications lead to wrong predictions. 

Popular methods for generating adversarial sequence use the gradient of the loss function to change the initial sequences. 
The first practical and fast algorithm based on the gradient method is the Fast Gradient Sign Method (FGSM)~\cite{goodfellow2014explaining}. 
One attack inspired by the FGSM is the high-performing BERT-attack~\cite{li2020bert}. 
The attack uses a pre-trained language model BERT to replace crucial tokens with semantically consistent ones.

The existence of adversarial attacks implies the necessity of developing defence methods.
~\cite{yuan2017defencesreview} provides an overview, for example. 
One of the most popular approaches for defence is an adversarial training advocated in~\cite{goodfellow2014explaining,huang2015learning}. 
The idea of adversarial training is to expand a training sample with adversarial examples equipped with correct labels and fine-tune a deep-learning model with an expanded training sample. 
Similarly, in~\cite{zeager2017cardfraud}, the authors improve a fraud detection system by adding fraudulent transactions in the model's training set. 
Another approach is adversarial detection. 
In this case, we train a separate detector model.
\cite{metzen2017detecting} describe the detector of adversarial examples as a supplementary network to the initial neural network. 
Meanwhile~\cite{feinman2017bayesian} provide a different approach to detecting adversarial examples based on Bayesian theory.  
They detect adversarial examples using a Bayesian neural network and estimating the uncertainty of the input data.

Adversarial attacks and defences from them are of crucial importance in financial applications~\cite{zeager2017adversarial}.
The literature of adversarial attacks on transaction records includes~\cite{fursov2020differ,fursov2020gradient}. 
However, the authors do not consider the peculiarities of transaction records data and apply general approaches to adversarial attacks on discrete sequence data.
Also, they pay little attention to efficient defences from such attacks and how realistically one can obtain such sequences.

There are no comprehensive investigations on the reliability of deep-learning models aimed at processing sequential data from transaction records in the literature.
It stems from the lack of attention to the transaction records data's peculiarities and the limited number of problems and datasets. 

This study fills this void by being the first encompassing work on adversarial attacks on transaction records data and corresponding defences for diverse datasets while taking data peculiarities into account.
According to our knowledge, no one before explored attacks with additions of tokens to the end and training specific generative model for financial transactions data that is a sequence of a mix of discrete and continuous variables.
Moreover, as the data is interpretable, it is vital to consider, what factors affect vulnerability of models to adversarial attacks.
 \vspace{-5mm} 
\section{Methods}
\label{sec:methods}
 \vspace{-3mm} 
\paragraph{Blackbox vs whitebox scenario}

We focus on the most realistic blackbox scenario~\cite{szegedy2013intriguing}.
The attack in this scenario does not have access to the original model.
Instead, the attacker can use scores for a substitute model trained with a different subset of data.
A substitute model can have different hyperparameters or even a different architecture replicating real-life situations. The attacker has little knowledge about the model's real architecture or parameters but can collect a training data sample to train a model. 
 \vspace{-5mm} 
\subsection{Attack methods}
\label{sec:attack_methods}
\begin{figure}[H]
 \centering
     \includegraphics[width=0.375\columnwidth]{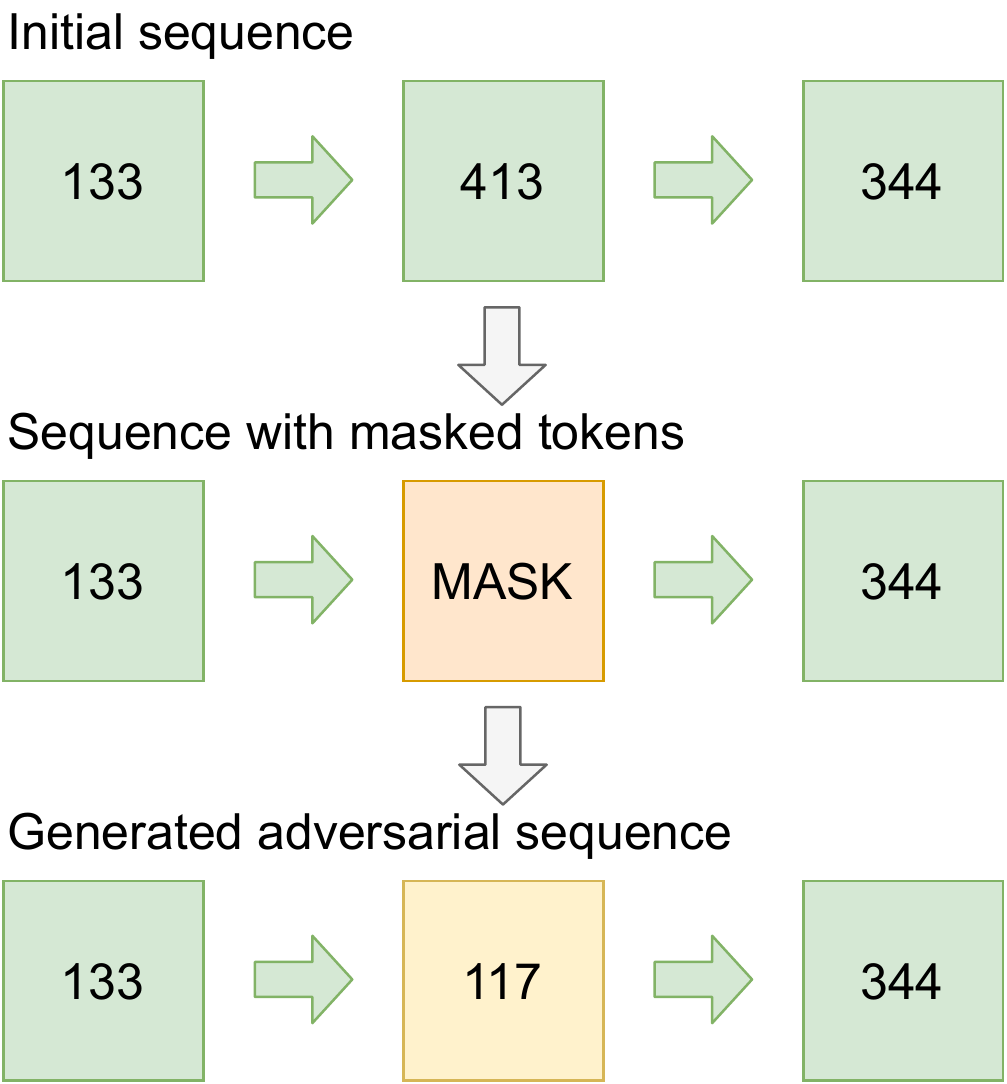}
     \caption{Example of an attack with token replacements where we select tokens to replace and then find the most effective tokens for a replacement to obtain an adversarial sequence}
    \label{fig:replace_attack_scheme}
\end{figure}

\begin{figure}[H]   
 \centering
 \includegraphics[width=0.5\columnwidth]{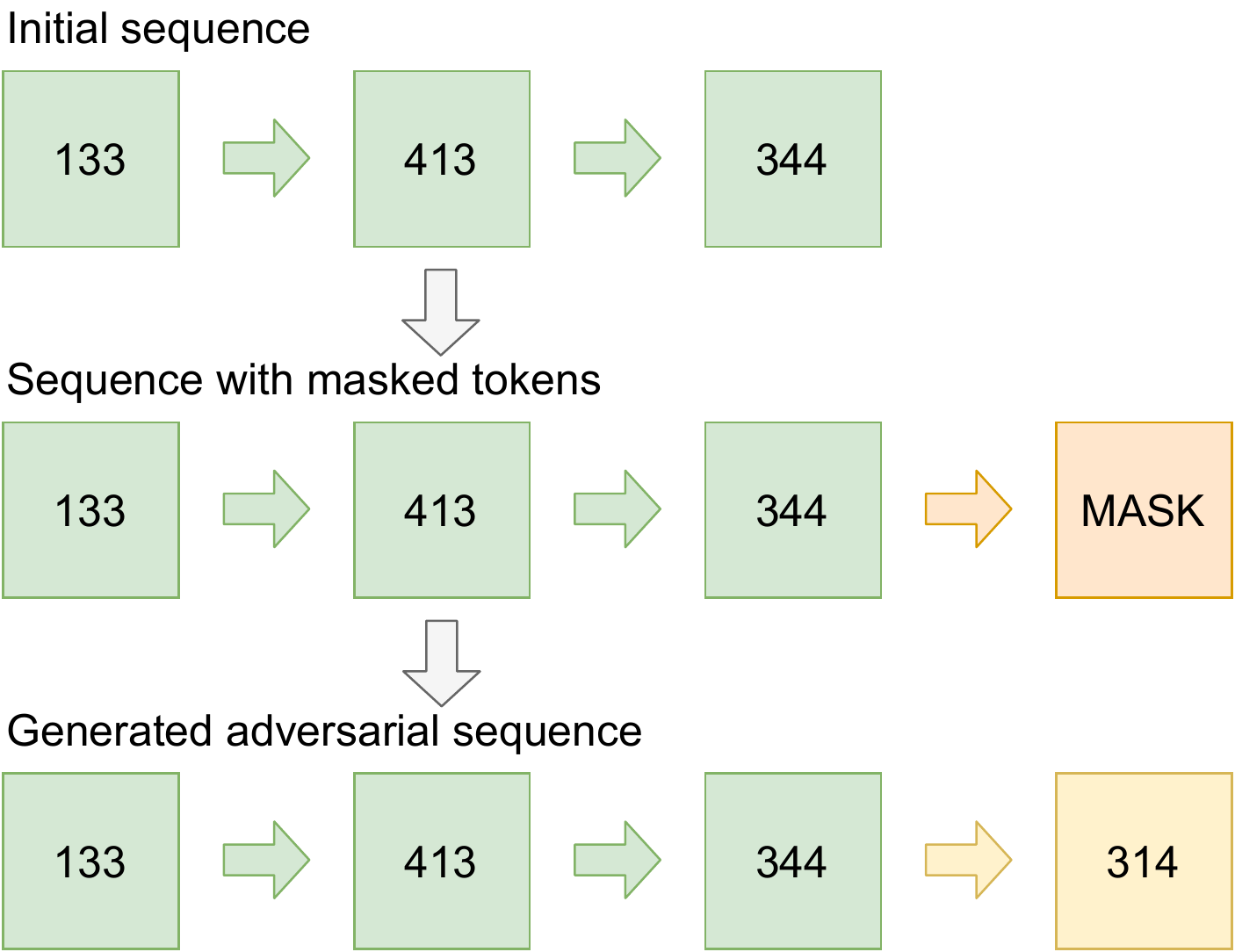}
     \caption{Example of an attack with token additions where we add the most adversarial tokens to the end of an adversarial sequence}
     \label{fig:concat_attack_scheme}
\end{figure}

 \vspace{-3mm} 
We consider realistic attack scenarios in the financial sector.
For this reason, our attacks resemble existing approaches in other communities.
We adapt them to the specificity of transaction records data. 
We categorize the algorithms by attack method into two types.
In the first case, an attack edits an original sequence of transactions by \textbf{replacing existing tokens}. 
We present the scheme of such attack in~\autoref{fig:replace_attack_scheme}.
For the second case, an attack \textbf{concatenates new adversarial transactions to the end} of the original sequence. 
For such attacks, we use the \emph{Concat} prefix in the name. 
The scheme of such attack is in~\autoref{fig:concat_attack_scheme}.


\paragraph{Replacement and concatenation attacks}




\textbf{Sampling Fool (SF)}~\cite{fursov2020gradient}. 
It uses a trained Masked Language Model (MLM)~\cite{devlin2018bert}, to sample from a categorical distribution new tokens to replace random masked tokens. 
Then it selects the most adversarial example among a fixed number of generated sequences. 
We generate $100$ sequences in our experiments.

\begin{figure}[H]
    \centering
    \includegraphics[width=0.5\columnwidth]{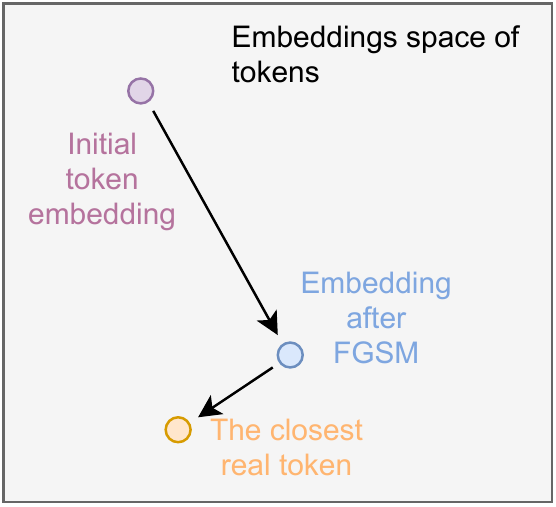}
    \caption{FGSM attack scheme in an embedded space of tokens}
    \label{fig:fgsm_scheme}
\end{figure}

\textbf{FGSM (Fast Gradient Sign Method)}~\cite{liang2017deep}. 
The attack selects a random token in a sequence and uses the Fast Gradient Sign Method to perturb its embedding vector. 
The algorithm then finds the closest vector in an embedding matrix and replaces the chosen token with the one that corresponds to the identified vector. 
We show the FGSM attack in an embedded space of tokens in~\autoref{fig:fgsm_scheme}. 
If we want to replace several tokens, we do it greedily by replacing one token at a time.
We select the amounts from a categorical distribution given the overall limit for these attacks if not specified otherwise. 

\begin{figure}[H]
    \centering
    \includegraphics[width=0.8\columnwidth]{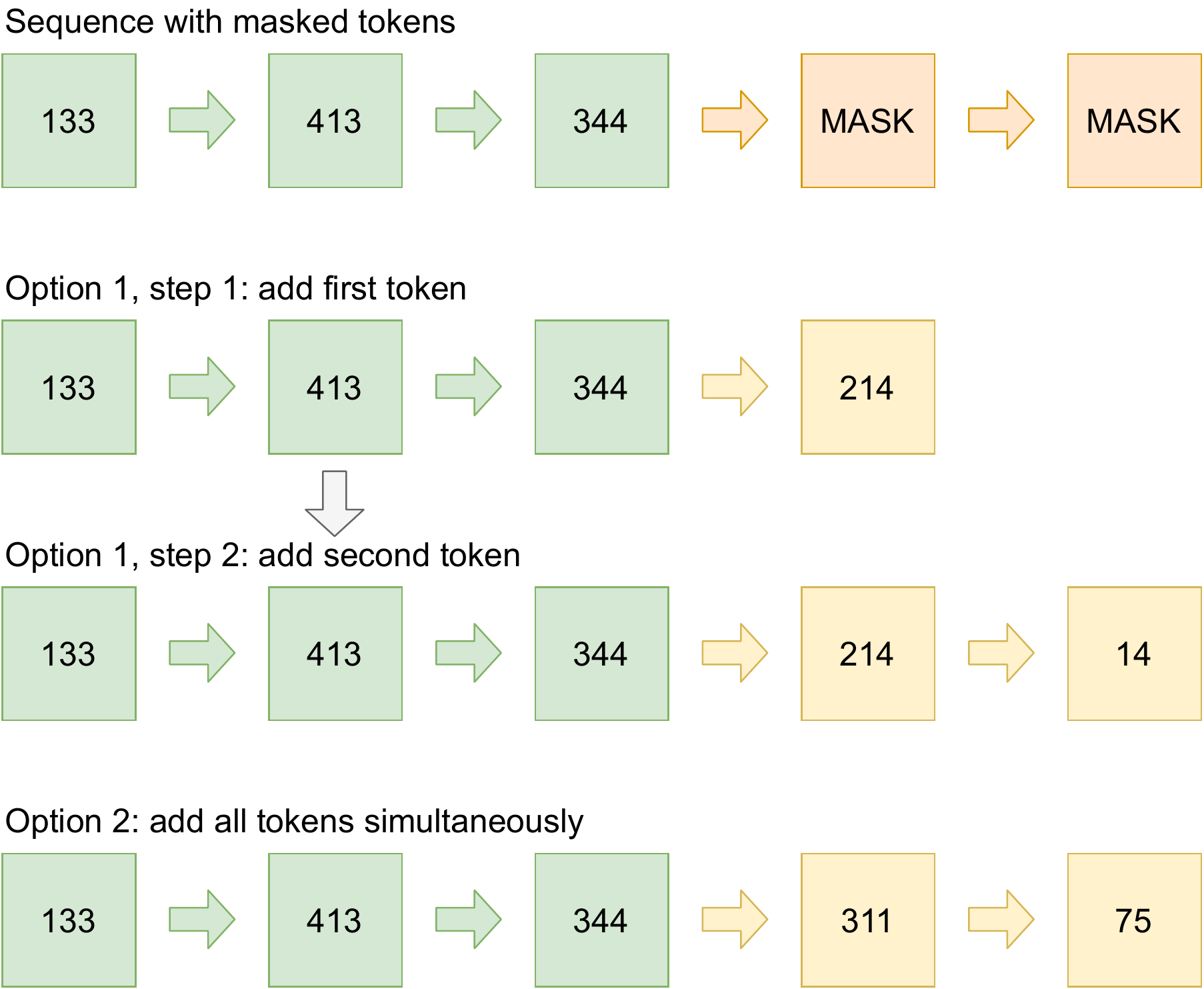}
    \caption{Difference between sequential (Option 1) and simultaneous (Option 2) Concat FGSM attacks}
    \label{fig:seq_sim_diff}
\end{figure}

\textbf{Concat FGSM}. 
This algorithm extends the FGSM idea. 
Here, we add $k$ random transactions to the original sequence and run the FGSM algorithm only on the added transactions. 
As a result, the edit-distance between the adversarial example and the original will be exactly $k$. 
We use two variations of this approach. 
They are Concat FGSM, [seq], and Concat FGSM, [sim]. 
In the first option, sequential, we add tokens one by one at the end of the sequence. 
In the second, simultaneous, we start by adding a random number of mask tokens and then replacing them by FGSM simultaneously taking into account the context. 
We show the difference between both approaches in~\autoref{fig:seq_sim_diff}. 

\textbf{LM FGSM}. 
To increase the plausibility of generated adversarial examples, LM FGSM uses a trained autoregressive language model to estimate which transaction is the most appropriate in a given context. 
The algorithm uses FGSM to find the closest token with the perplexity less than a predefined threshold $\tau$. 
Thus, LM FGSM increase the plausibility of generated adversarial examples.
A similar idea appears in~\cite{ebrahimi2018hotflip} for NLP data.

\textbf{Concat Sampling Fool (Concat SF)}. 
This method replicates the idea of Concat FGSM and adds $k$ random transaction tokens to the original sequence. 
Then, using the pre-trained BERT language model, we receive the vocabulary probabilities for each added token. 
As a result, the algorithm samples from the categorical distribution and chooses the best adversarial example using a given temperature value.

\textbf{Sequential Concat Sampling Fool (Seq Concat SF)}. 
The difference between this algorithm and the usual Concat SF is that this algorithm, instead of immediately adding $k$ tokens to the original sequence, adds tokens one at a time. 
We choose the token at each step that reduces the model's probability score for the target (correct) class the most. 

\vspace{-5mm} 
\subsection{Defense methods}
\vspace{-3mm} 
We use two approaches to defend from adversarial attacks with ideas from the general defences literature~\cite{wang2019survey}.   
They are adversarial training and adversarial detection.

\textbf{Adversarial training}. 
The adversarial training objective is to increase the model robustness by adding adversarial examples into the training set. 
The most popular and common approach is adding the adversarial examples with correct labels to the training set, so the trained model correctly predicts the label of future adversarial examples after fine-tuning. 
Other popular adversarial training methods generate adversarial examples at each training iteration~\cite{madry2017towards}. From the obtained examples, we calculate the gradients of the model and do a backpropagation.

\textbf{Adversarial detection}. 
This method consists of training an additional model, a discriminator, which solves a binary classification model addressing whether a present sequence is real or generated by an adversarial attack. 
The trained discriminator model can detect adversarial examples and prevent possible harm from them.
\vspace{-5mm} 
\section{Data}
\vspace{-3mm} 
We consider sequences of transaction records stemming from clients at financial institutions.
For each client, we have a sequence of transactions. 
Each transaction is a triplet consisting of three elements. 
The first one is a discrete code to identify the type of transaction (Merchant Category Code). 
Among others, it can be a drug store, ATM or grocery store.
The second corresponds to a transaction amount, and the third is a timestamp for the transaction time.
For the transaction codes and the formatting of the data, we take~\cite{babaev2020metric} as an inspiration.

Moreover, we consider diverse real-world datasets for our numerical experiments taken from three major financial institutions.
In all cases, the input consists of transactions from individual clients. 
For each client, we have an ordered sequence of transactions. 
Each entry contains a category identifier such as a bookstore, ATM, drugstore, and more as well as a monetary value.
Further, the first and second sample's target output, \emph{Age 1} and \emph{Age 2}, is the client's age bin with four different bins in total.
For the third sample, \emph{Leaving}, the goal is to predict a label corresponding to a potential customer churn.
The fourth sample, \emph{Scoring}, contains a client default for consumer credit as a target.
We present the main statistics for both the data and the models we use in~\autoref{tab:datasets}.
For our critical experiments, we provide results for all datasets. In other cases, due to space constraints, we focus on one or two datasets, as results are similar if not specified otherwise.

We train the target model on $65\%$ of the original data. 
Similarly, we use for the substitute model the rest of the data, $35\%$.
Additional implementation details are in Appendix.

\begin{table*}[t]
    \centering
    \begin{tabular}{ccccccc}
    \hline
    \multicolumn{2}{c}{Dataset}	& Age 1  & Age 2 & Client leaving & Scoring \\
    \hline
 \multicolumn{2}{c}{$\#$classes}	& 4	& 4 & 2 & 2 \\
\multicolumn{2}{c}{Mean sequence length}	& 86.0	& 42.5 & 48.4 & 268.0 \\
\multicolumn{2}{c}{Max sequence length}	& 148 & 148  & 148 & 800 \\
\multicolumn{2}{c}{Number of transactions}	& 2 943 885 & 3 951 115 & 490 513 & 3 117 026 \\
\multicolumn{2}{c}{Number of clients} & 29 973  & 6886  & 5000 & 332 216 \\
\multicolumn{2}{c}{Number of code tokens} &	187	 & 409 & 344 & 400 \\
\hline
\multicolumn{6}{c}{Models' accuracy for the test data} \\
\hline
& Substitute & 0.537  & 0.675  & 0.653 & 0.825\\
\multirow{-2}{*}{LSTM} & Target     & 0.551     & 0.668 & 0.663  & 0.881 \\ \hline
& Substitute & 0.548 & 0.675 & 0.644 & 0.833\\ 
\multirow{-2}{*}{GRU}  & Target    & 0.562  &   0.663 & 0.672 & 0.862 \\ \hline
& Substitute &    0.553  & 0.674 & 0.660 & 0.872 \\ 
\multirow{-2}{*}{CNN}  & Target    &  0.564  &  0.674 & 0.670 &  0.903 \\ \hline

    \end{tabular}
    \caption{Datasets statistics and models' accuracy for the test data}
    \label{tab:datasets}
\end{table*}


\vspace{-5mm} 
\section{Experiments}
\vspace{-3mm} 
We provide the code, experiments, running examples and links to the transaction datasets on a public online repository\footnote{\url{https://github.com/fursovia/adversarial_sber}}.
\vspace{-4mm} 
\subsection{Metrics}
\vspace{-3mm} 
To measure the quality of attacks, we find an adversarial example for each example from the test sample. We use the obtained adversarial examples to calculate a set of metrics and assess the attacks' quality.
We then average over all considered examples.
As we compare multiple adversarial attacks, we provide a unified notation to facilitate their understanding. We consider the following elements in our notation: 
\begin{itemize}
    \item $C^{t}$ --- The target classifier that we want to "deceive" during the attack;
    \item $\mathbf{x}_{i}$ --- An original example from the test sample.
    \item $\tilde{\mathbf{x}}_{i}$ --- An adversarial example that we obtain during the attack for an example~$\mathbf{x}_{i}$.
    \item $C^{t}(\mathbf{x}_{i})$ -- The prediction of the original classifier for the object $\mathbf{x}_{i}$, a class label.
    \item $C^{t} (\tilde{\mathbf{x}}_{i})$ --  The prediction of the original classifier for the adversarial example $\tilde{\mathbf{x}}_{i}$, a class label.
\end{itemize} 

With this notation, we can define a list of four metrics to evaluate our data against them.
\begin{enumerate}
    \item \textbf{Word Error Rate (WER)}.
Before we can create an adversarial attack, we must change the initial sequence. 
Our change can be either by inserting, deleting, or replacing a token in some position in the original sequence. 
In the $WER$ calculation, we treat any change to the sequence made by insertion, deletion, or replacement as $1$. 
Therefore, we consider the adversarial sequence perfect if $WER = 1$ and the target classifier output has changed.
    
  \item \textbf{Adversarial Accuracy (AA)}.
AA is the rate of examples that the target model classifies correctly after an adversarial attack.
We use the following formula to calculate it:
$
    \mathrm{AA}(A) = \frac{1}{N} \sum_{i=1}^{N} \mathbb{1} \{C^{t}(\mathbf{x}_{i}) = C^{t} (\tilde{\mathbf{x}}_{i}) \} 
    $.
    
    \item \textbf{Probability Difference (PD)}. 
The PD demonstrates how our model's confidence in the response changes after an adversarial attack:
    $
    \mathrm{PD}(A) = \frac{1}{N} \sum_{i=1}^{N} C_{\mathbb{P}}^{t}(\mathbf{x}_{i}) - C_{\mathbb{P}}^{t} (\tilde{\mathbf{x}}_{i}),
    $
where $C_{\mathbb{P}}^{t}(\mathbf{x}_{i})$ is the confidence in the response for the original example $\mathbf{x}_{i}$, and $C_{\mathbb{P}}^{t} (\tilde{\mathbf{x}}_{i})$ is the confidence in the response for the adversarial example $\tilde{\mathbf{x}}_{i}$.

    \item \textbf{NAD (Normalized Accuracy Drop)}
NAD is a probability drop, normalized on WER.
For a classification task, we calculate the Normalized Accuracy Drop in the following way:
    \[
    \mathrm{NAD}(A) = \frac{1}{N} \sum_{i=1}^{N} \frac{\mathbb{1} \{C^{t}(\mathbf{x}_{i}) \not= C^{t} (\tilde{\mathbf{x}}_{i}) \} }{\mathrm{WER}(\mathbf{x}_{i}, \tilde{\mathbf{x}}_{i})},
    \]

NAD reflects a trade-off between the number of changed tokens and model quality on adversarial examples.
\end{enumerate}
\vspace{-4mm} 
\subsection{Overall attack quality}
\label{subsec:overall_quality}
\vspace{-2mm} 
We compare the performance of six attacks. 
They are Sampling Fool (SF), Concat Sampling Fool (Concact SF), Sequential Concat Sampling Fool (Seq Concat SF), FGSM, Concat FGSM, Concat LM FGSM and we describe them in ~\autoref{sec:attack_methods}. 
The results are in~\autoref{tab:general_results} for the GRU target and the substitute model architectures. 
Similarly, in~\autoref{tab:white_general_results}, we have them for the sake of comparison for the white-box scenario, when we use the target model as the substitute for the most promising and representative attacks.

We observe that all adversarial attacks have a high quality. 
After we change on average no more than $2$ tokens, we can fool a model with a probability of at least $0.8$.
This probability makes the attacked model useless, while the number of added or modified transactions is small compared to the total number of transactions in the sequence judging by Word Error Rate. 
We can completely break the model for some attacks by adding only one token at the end.
Attacks, where we add tokens at the end, perform comparably to attacks, where we include tokens in the middle of a sequence. 
So, these more realistic attacks are also powerful for money transactions data.
The quality of attacks is comparable to the quality of the Greedy baseline attack.
It is the Greedy attack based on a brute force selection of tokens for insertion or editing. 
Thus, the attack provides close to the best achievable performance in our black-box scenario.
However, FGSM-based attacks provide better performance scores than SamplingFool-based ones due to the random search for adversarial examples in the second case. It can be useful to unite these approaches to create a generative model that can generate sequences of transaction records that are both realistic and adversarial~\cite{fursov2020differ}.
Also, more realistic concatenation of tokens to the end of a sequence results in lower performance scores. 
In sections below we consider the most successful and the most representative attacks SF, Concat SF, FGSM, and Concat FGSM.

\begin{table*}[t]\parbox{.45\linewidth}{
\begin{tabular}{ccccc}
\hline
Attack & NAD $\uparrow$ & WER $\downarrow$ & AA $\downarrow$  & PD $\uparrow$ \\
\hline
\multicolumn{5}{c}{Age 1 (Accuracy 0.562)} \\
\hline
SF & 0.09 & 12.87 & \textbf{0.45} & 0.07\\
Concat SF & 0.26 & 2 & 0.49 & 0.06\\
Seq Concat SF & 0.28 & 2 & \textbf{0.45} & 0.09\\
FGSM & \textbf{0.45} & 1.79 & 0.49 & 0.04\\
Concat FGSM & 0.13 & 4 & 0.47 & 0.07\\
LM FGSM & 0.44 & \textbf{1.78} & 0.50 & 0.04\\

\hline
\multicolumn{5}{c}{Age 2 (Accuracy 0.663)} \\ \hline
SF & 0.45 & 5.18 & 0.29 & 0.07 \\
Concat SF & 0.38  & \textbf{2} & 0.25 & 0.07\\
Seq Concat SF & 0.39 & \textbf{2} & \textbf{0.12} & 0.09\\
FGSM & \textbf{0.71} & 3.46 & 0.21 & 0.08\\
Concat FGSM & 0.2 & 4.0 & 0.19 & 0.07\\
LM FGSM & \textbf{0.71} & 3.45 & 0.20 & 0.08 \\
\hline

\multicolumn{5}{c}{Client leaving (Accuracy 0.672)} \\ \hline
SF & 0.14 & 14.78 & 0.69 & 0.00\\
Concat SF & 0.19 & \textbf{2} & 0.62 & 0.00\\
Seq Concat SF & 0.23 & \textbf{2} & 0.55 & 0.03 \\
FGSM & 0.23 & 8.69  & 0.47 & 0.10\\
Concat FGSM & 0.14 & 4 & 0.46 & 0.09\\
LM FGSM & \textbf{0.24} & 8.84 & \textbf{0.44} & 0.11 \\
\hline
\multicolumn{5}{c}{Scoring (Accuracy 0.86)} \\
\hline
SF & 0.15 & \textbf{5.74} & 0.78 & 0.00 \\
Concat SF & 0.05 & 6 & 0.73 & 0.05 \\
FGSM & \textbf{0.27} & 8.43 & 0.67 & 0.10 \\
Concat FGSM & 0.10 & 6 & \textbf{0.44} & 0.22 \\
LM FGSM & \textbf{0.27} & 7.76 & 0.67 & 0.09\\
\hline
    \end{tabular}
    \caption{
        Summary of the effectiveness of black-box attacks with GRU as the target and the substitute models architectures for different datasets with initial accuracy values in brackets. $\uparrow$ marks metrics we want to maximize, $\downarrow$ marks metrics we want to minimize.}
    \label{tab:general_results}
}
\hfill \parbox{.45\linewidth}{

\begin{tabular}{ccccc}
\hline
Attack & NAD $\uparrow$ & WER $\downarrow$ & AA $\downarrow$  & PD $\uparrow$ \\
\hline
\multicolumn{5}{c}{Age 1 } \\
\hline
SF & 0.13 & 13.18 & 0.07 & 0.23\\
Concat SF & 0.38 & \textbf{2} & 0.23 & 0.24\\
FGSM & \textbf{0.70} & 2.51 & \textbf{0.00} & 0.29\\
Concat FGSM & 0.25 & 4  & \textbf{0.00} & 0.48\\
LM FGSM & 0.68 & 2.68 & 0.01 & 0.27 \\
\hline
\multicolumn{5}{c}{Age 2 } \\ \hline
SF & 0.56 & 4.81 & 0.22 & 0.09\\
Concat SF & 0.42 & \textbf{2} & 0.15 & 0.12\\
FGSM & 0.79 & 2.40 & \textbf{0.01} & 0.15\\
Concat FGSM & 0.24 & 4 & 0.05 & 0.25\\
LM FGSM & \textbf{0.84} & 2.13 & 0.03 & 0.14 \\
\hline

\multicolumn{5}{c}{Client leaving } \\ \hline
SF & 0.29  & 14.37 & 0.50 & 0.09\\
Concat SF & 0.23 & \textbf{2} & 0.53 & 0.06\\
FGSM & \textbf{0.42} & 6.51 & \textbf{0.09} & 0.15\\
Concat FGSM & 0.16 & 4  & 0.35 & 0.18\\
LM FGSM & \textbf{0.42} & 6.37 & 0.11 & 0.15 \\
\hline
\multicolumn{5}{c}{Scoring} \\
\hline
SF & 0.15 & \textbf{5.74} & 0.79 & 0.15 \\
Concat SF & 0.03 & 6 & 0.80 & 0.22 \\
FGSM & \textbf{0.47} & 8.43 & \textbf{0.02} & 0.28 \\
Concat FGSM & 0.14 & 6 & 0.14 & 0.43 \\
LM FGSM & 0.46 & 7.76 & 0.04 & 0.28 \\
\hline
    \end{tabular}
    \caption{Summary of the effectiveness of the most promising white-box attacks with GRU as the target model architecture}
    \label{tab:white_general_results}}
\end{table*}
\vspace{-4mm} 
\subsection{Dependence on the architecture}
\vspace{-2mm} 
A malicious user can lack knowledge regarding the specific architecture used for data processing. 
Hence, we want to address how the attack's quality changes if the architectures of both the attacked target and the substitute model used for the attack differ.
For this, we present the results when the attack targets an LSTM model and a CNN the substitute model in~\autoref{tab:lstm_vs_cnn}.
Comparing these results to the previous section results, we see that the attack's quality in the case of models of different nature deteriorates markedly. 
However, in both cases, the Concat FGSM attack works reasonably by adding strongly adversarial tokens at the end of the sequence. 
We conjecture that this is because RNN-based models pay more attention to the tokens in the end, while a CNN model is position-agnostic regarding a particular token. 
So, Concat FGSM attacks are successful, even if a substitute model's architecture is different from that of a target model.

\begin{table*}[t]
    \centering
\begin{tabular}{ccccc}
\hline
Attack & NAD $\uparrow$ & WER $\downarrow$ & AA $\downarrow$  & PD $\uparrow$ \\
\hline
\multicolumn{5}{c}{Age 1} \\
\hline
SF & 0.10 & 13.75& 0.45 &  0.12\\
Concat SF  & 0.30 & 2& 0.40&0.14 \\
FGSM & 0.40 &3.81& 0.48 & 0.10\\
Concat FGSM & 0.18 & 4 & 0.30 & 0.20 \\
LM FGSM & 0.43 & 3.53 & 0.45 & 0.10 \\
\hline
\multicolumn{5}{c}{Age 2} \\
\hline
SF &  0.21 & 14.92& 0.78 &  -0.10\\
Concat SF  & 0.09 & 2& 0.82 & -0.81 \\
FGSM & 0.21 & 8.90 &0.78 &  -0.13 \\
Concat FGSM & 0.05 & 4 & 0.80 & -0.10\\
LM FGSM & 0.58 & 3.37 & 0.38 &  0.02\\
\hline
\multicolumn{5}{c}{Client leaving} \\
\hline 
SF & 0.09 & 12.17& 0.46 & 0.03 \\
Concat SF  & 0.25 & 2 &0.50 & 0.04 \\
FGSM & 0.46 & 6.67 & 0.30 &  0.10 \\
Concat FGSM & 0.14 & 4 & 0.43 & 0.08 \\
LM FGSM & 0.21 & 12.53 & 0.55 & 0.07\\
\hline
\multicolumn{5}{c}{Scoring} \\
\hline 
SF & 0.10 & 6.31 & 0.84 & 0.05 \\
Concat SF & 0.06 & 6 & 0.65 & 0.18 \\
FGSM & 0.17 & 13.92 & 0.76 & 0.12\\
Concat FGSM & 0.06 & 6 & 0.64 & 0.18 \\
\hline
\end{tabular}
    \caption{Black-box attack effectiveness with LSTM as the target model and CNN as the substitute model}
    \label{tab:lstm_vs_cnn}
\end{table*}

\vspace{-5mm} 
\subsection{Defenses from adversarial attacks}
\label{subsec:defenses_results}
\vspace{-1mm} 
\subsubsection{Adversarial training}
\begin{figure}[H]
    \centering
    \includegraphics[scale=0.23]{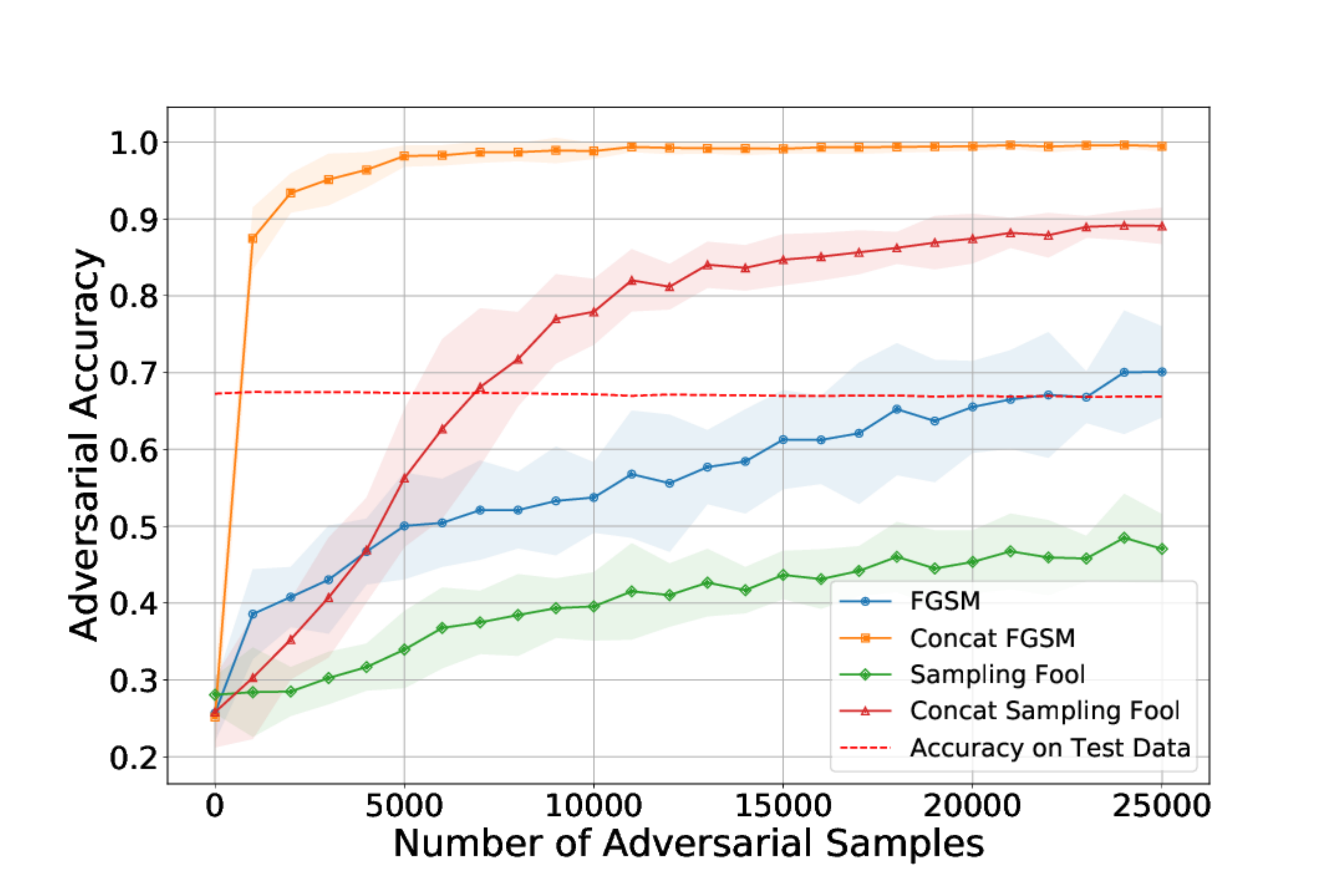}
    \caption{Adversarial Accuracy metric vs number of adversarial examples added to the training set for dataset Age 2.}
    \label{fig:adversarial_training_age_1}
\end{figure}

\vspace{-1mm} 
\autoref{fig:adversarial_training_age_1} presents the adversarial training results for one dataset. The target and substitute models are GRU. 
In the figure, we average the results from $10$ runs and present mean values as solid curves and a mean $\pm$ with two standard deviations as a shaded area. 
For most attacks, the Adversarial Accuracy quickly increases when we compare it against the overall accuracy presented as the dashed line.
The results for other datasets are similar.
For most attacks, we see that it is enough to add about $15000$ adversarial examples to the training sample and fine-tune model using an expanded training sample to make model robust against a particular adversarial attack.
However, after adversarial training, a SamplingFool attack is possible despite being worse on our attack quality metrics.

\vspace{-5mm} 
\subsubsection{Adversarial detection}
\vspace{-3mm} 
\begin{table*}[t]
\centering
\begin{tabular}{cccccc}
\hline

Attack &  Age 1  & Age 2 & Client  & Scoring \\
       &         &       & leaving & \\
\hline
SF & \textbf{0.500}  & \textbf{0.623} & \textbf{0.661} & \textbf{0.422} \\
Concat SF & 0.998  & 0.986 & 0.967 & 0.988 \\
FGSM  & 0.493  & 0.946 & 0.800 & 0.953 \\
Concat FGSM & 1.000  & 0.995 & 0.991 & 1.000 \\

\hline



\end{tabular}
\caption{Accuracy of adversarial examples detection for different types of attacks and datasets}
\label{tab:advers_detection}
\end{table*}

Another approach for defending against adversarial attacks is the adversarial sequences detection.
Carrying out this type of attack entails that we train a separate classifier detecting a sequence from an original data distribution or stemming from an adversarial attack.
If this classifier is of high quality, we can easily protect our model from adversarial attacks by discarding potentially adversarial sequences.

As the detector classifier, we use a GRU model. 
We train it for $10$ epochs. 
We notice that increasing the number of epochs does not lead to improving the detector's quality. 

The classification data is balanced, so we calculate the accuracy for evaluating the quality of adversarial detection. 
These results for different attacks are in~\autoref{tab:advers_detection}.

We can detect with accuracy greater than $0.9$ the adversarial examples generated by most of the attacks.
However, adversarial sequences from the Sampling Fool attack are more difficult to detect.
Therefore, we postulate that we can effectively repel many attacks with the help of a trained detector.
\vspace{-4mm} 
\subsection{Dependence on additional amount}
\vspace{-2mm} 
In addition to a discrete token, each transaction has an associated money amount. 
A strong dependence on an attack's performance based on using additional monetary amounts can be a show stopper, as a malicious user tries to minimize the amount of money spent on additional transactions. 

We experiment on the Age 1 dataset by measuring the Concat FGSM attack's quality with varying constraints starting from few constraints and going to the case where there is practically no limit. 

In~\autoref{tab:concat_fgsm_sums}, we present the results.
For this table, we consider the Age 1 dataset and add one token at the end of a transaction sequence, so WER equals $1$ for all attacks.
They demonstrate no difference in the attack's quality concerning the considered amount. 
Therefore, we can fool the model using a quite limited amount of monetary funds.
We elucidate that the model takes most of its information from the transaction token. 
As a result, attack strategies should focus on it.

\begin{table*}[t]
\centering
\begin{tabular}{cccc}
\hline
& Amount limit & AA &  PD\\
\hline
& 300 & 0.09 & 0.23 \\
& 500 & 0.12 & 0.20 \\
& 1000  & 0.2 & 0.17 \\
& 3000  & 0.05 & 0.27 \\
& 5000  & 0.01 & 0.31 \\
& 10000  & 0.01 & 0.35 \\
& 100000  & 0.04 & 0.31 \\
\hline
\end{tabular}

\caption{Overview of the Concat FGSM attack quality for different amounts of tokens for Age 1 dataset}
\label{tab:concat_fgsm_sums}
\end{table*}
\vspace{-4mm} 
\subsection{Closer look at generated sequences}
\vspace{-2mm} 
\label{subsec:closer_look}

Natural language data allow for a straightforward interpretation by experts.
A human can assess how realistic an adversarial sentence is.
For transaction records data, we can manually inspect any generated adversarial sequences and verify their realism level. 


     
\begin{figure}[H]
    \begin{center}
    \centerline{\includegraphics[width=0.9\columnwidth]{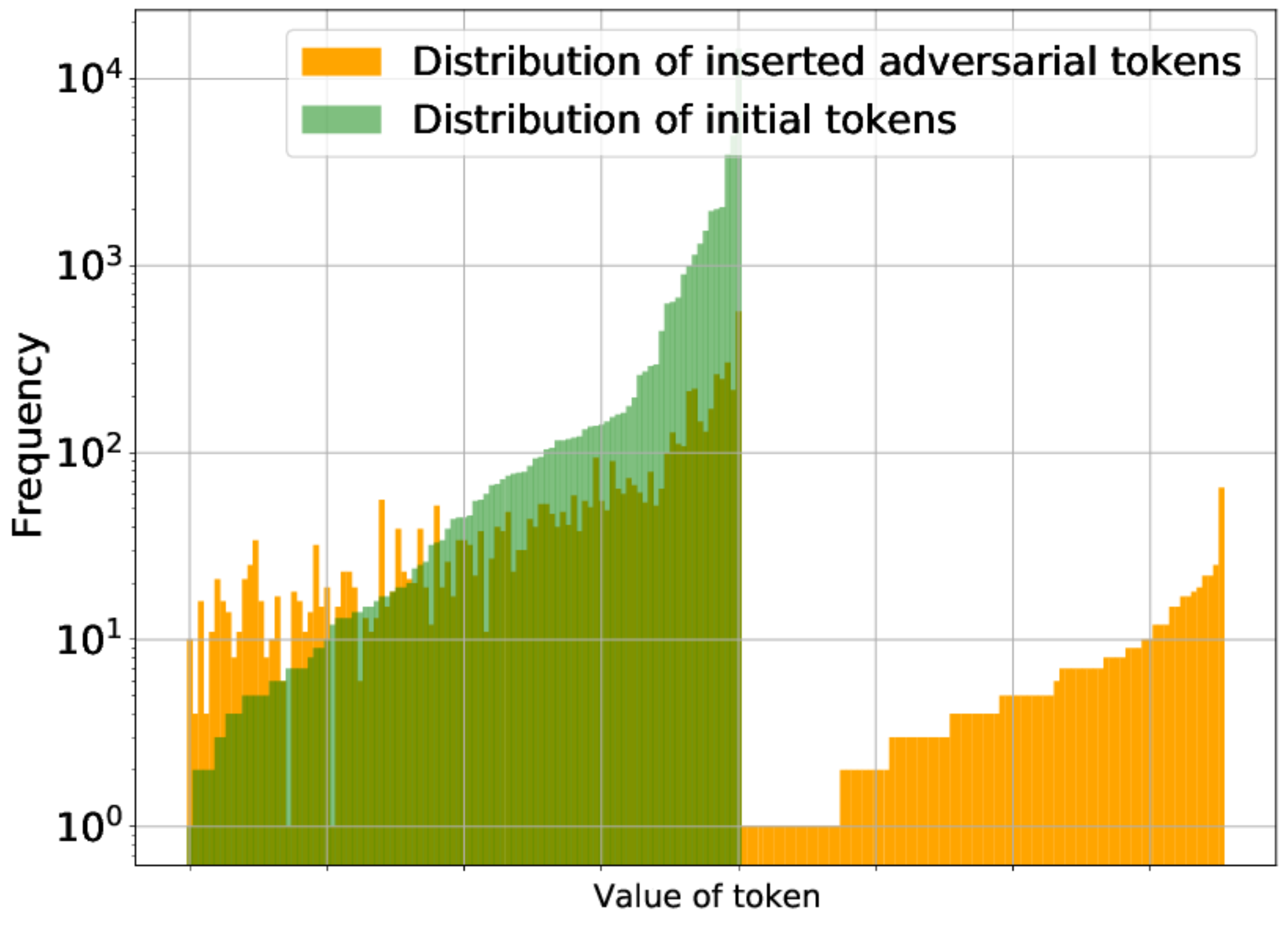}}
    \caption{The histograms of tokens from the initial transactions are in green and tokens inserted by the Sampling Fool attack in orange for the Age 1 dataset}
    \label{fig:age_sf}
    \end{center}
\end{figure}

The histograms of the initial distribution of tokens and tokens inserted by the Sampling Fool attack are in~\autoref{fig:age_sf}. 
We notice that most of the inserted tokens occur in the original sequences with similar frequencies, so the generated sequences are realistic from this point of view. 
However, some of the inserted tokens do not belong to the history of the client's transactions. 
Nevertheless, they are in the training set of the model.
This constellation also occurs in the case of the histogram of the Concat Sampling Fool.

\begin{figure}[H]
    \begin{center}
    \centerline{\includegraphics[width=0.9\columnwidth]{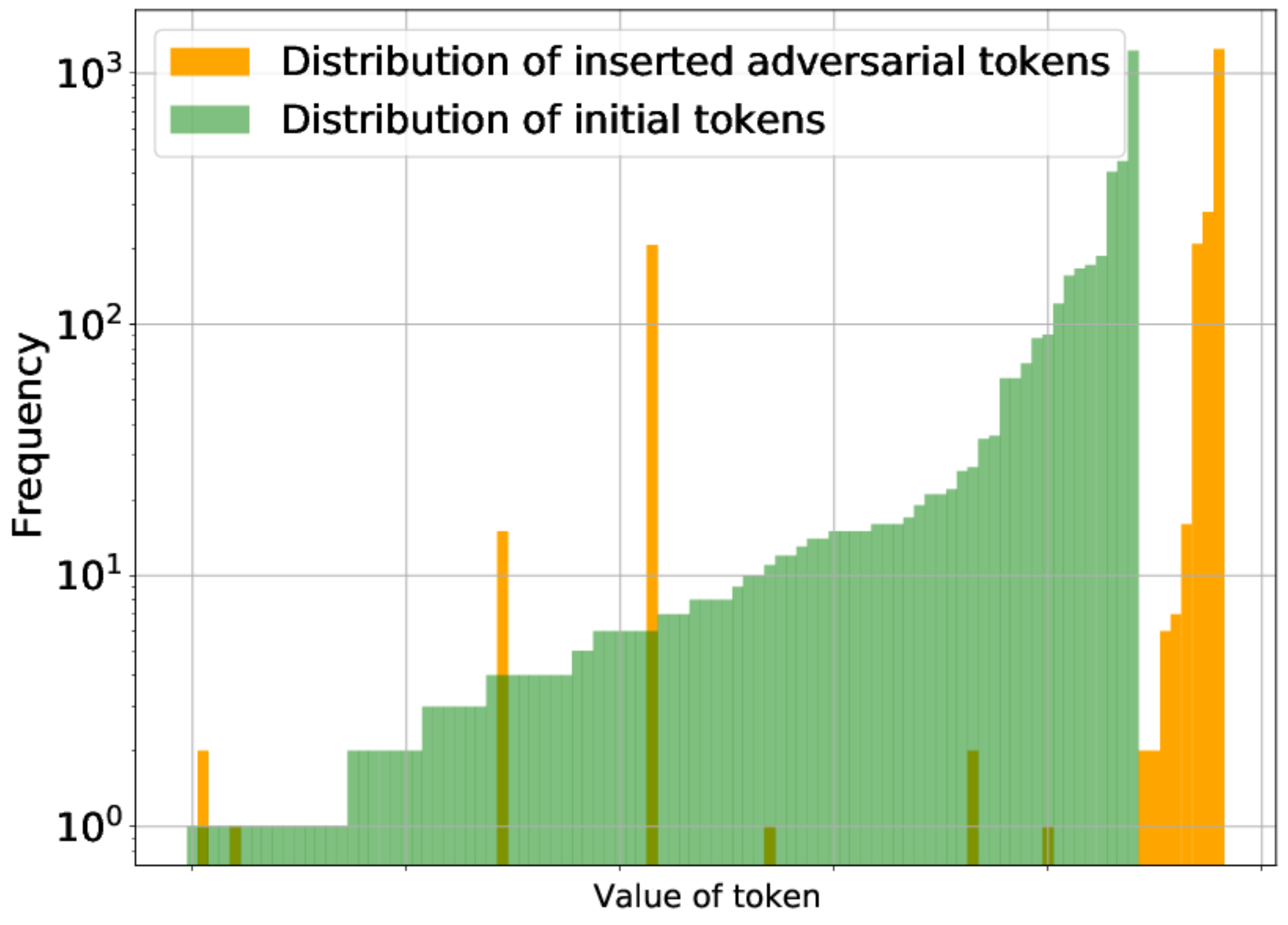}}
    \caption{The histograms of tokens from the initial transactions are in green and tokens inserted by Concat FGSM attack in orange for the Age 1 dataset}
    \label{fig:age_short_concat_fgsm}
    \end{center}
\end{figure}

We can observe different types of distributions for the Concat FGSM in~\autoref{fig:age_short_concat_fgsm}.
For this attack, only a limited number of tokens has a significant effect on the model prediction. 
The same phenomenon happens for the cases of the FGSM attacks.
So, we expect, that while these attacks are adequate given our metrics, it remains easy to detect them or fine-tune the model to make it resistant to this type of attacks.

\begin{figure}[H]
    \begin{center}
    \centerline{\includegraphics[width=0.8\columnwidth]{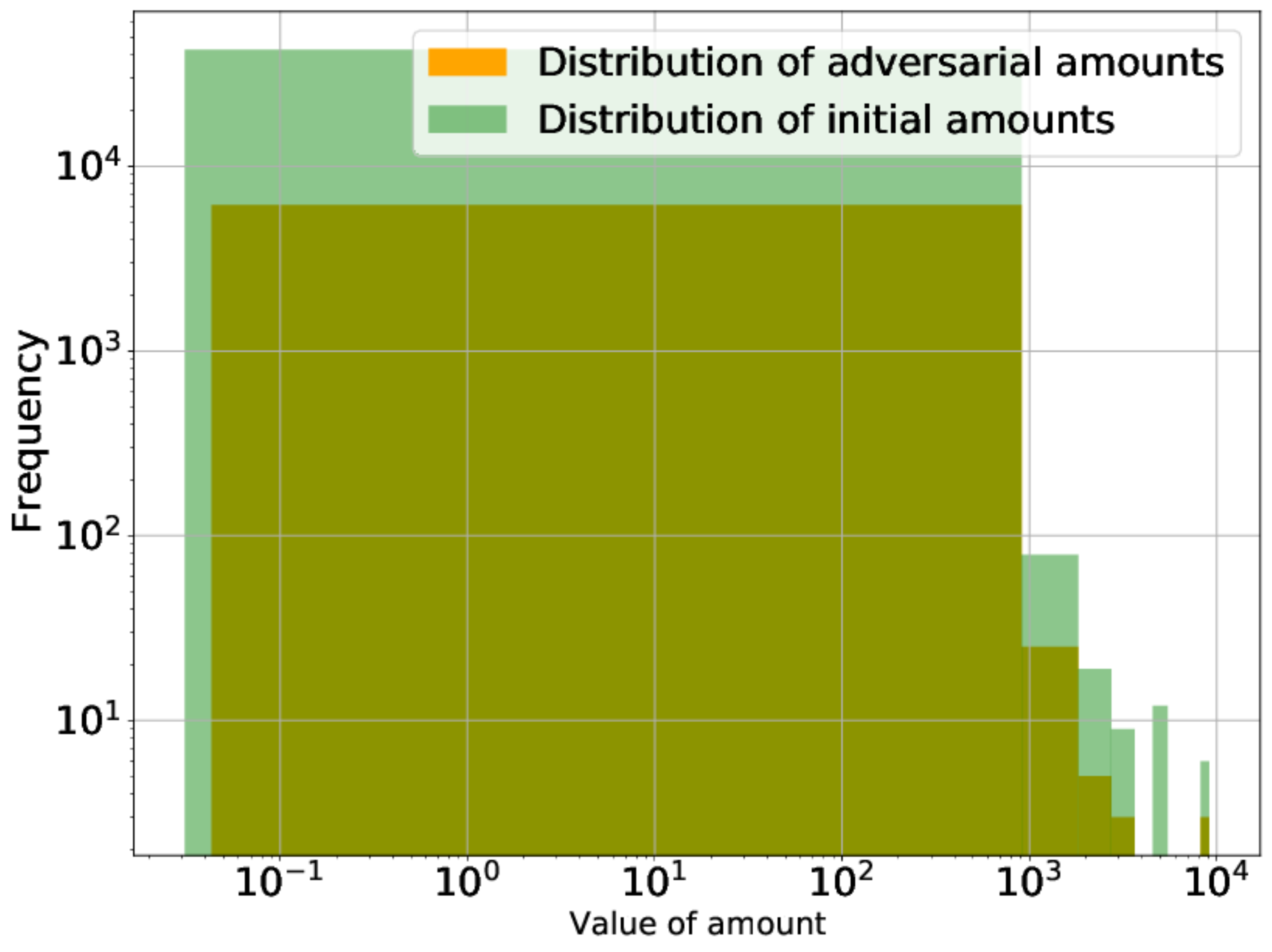}}
    \caption{The histograms of amounts for original transactions and amounts for transactions inserted by the Sampling Fool attack for a balanced Age 1 dataset of adversarial and initial normal sequences}
    \label{fig:amounts_age_sf}
    \end{center}
\end{figure}

We can summarize numerically the diversity of tokens added to the sequences through the considered attacks.
Three of our metrics are a diversity rate, a repetition rate, and a perplexity. 
For the first case, \textbf{the diversity rate} represents the number of unique adversarial tokens divided by the size of the vocabulary from which we sample these tokens. 
High diversity values suggest that an attacker uses a vast number of unique tokens and the generated adversarial sequences are harder to distinguish from sequences present in the original data.
In contrast, low diversity values suggest that an attack inserts the limited number of specific tokens.
For the second case, \textbf{the repetition rate} equals the number of tokens added by an attack appearing in an original sequence divided by the total number of added adversarial tokens.
High repetition rate means that an attack adds tokens that are already present in a sequence, thus the generated sequence is more realistic, while low repetition rate means that an attacker tries to insert new types of tokens, and that can be unrealistic.
Finally, \textbf{Perplexity} is the inverse normalized probability density that the specified sequence $\mathbf{x} = (x_1, x_2, \ldots, x_n)$ is derived from the specified distribution.
We define it as $ \mathrm{PP}(\mathbf{x}) = \mathbb{P} (x_{1}, x_2, \cdots, x_{n} ) ^{-\frac{1}{n}} = \sqrt[n]{\frac{1}{\mathbb{P} (x_1, x_2, \ldots, x_n) }}.$
We calculate the density using a trained language model. 
The lower the perplexity, the more plausible the sequence is from the language model's point of view.
\\ We observe the obtained diversity rate, repetition rate, and perplexity in~\autoref{tab:attacks_properties} for Age 1 dataset. For other datasets, the results are similar.
As the baselines we perform uniform sampling of tokens from the training set (Unif. Rand.) and sampling w.r.t. their frequencies of the occurrence (Distr. Rand.).
We see that SF and Concat SF's attacks for the language model generate more realistic adversarial sequences based on the considered criterion, while FGSM-based approaches and the Greedy attack do not pay attention to how realistic the generated sequences are. \\
\begin{table*}[t]
\centering
\begin{tabular}{cccccc}
\hline
Data & Attack &  Diversity $\uparrow$ & Repetition $\uparrow$ & Perplexity $\downarrow$ \\
& & rate & rate & \\
\hline
Age 1 
& Unif. Rand. & 0.99 & 0.47 & 7.49  \\
& Distr. Rand. & 0.37 & 0.99 & 3.02  \\
& SF  & 0.62 & \textbf{0.80} & \textbf{4.28} \\
& Concat SF & \textbf{0.77} & 0.58 & 6.01 \\
& FGSM & 0.15 & 0.57 & 4.71 \\
& Concat FGSM  & 0.08 & 0.11 & 17.76 \\
\hline
\end{tabular}
\caption{Diversity rate, repetition rates and perplexity of the adversarial sequences generated by the attacks}
\label{tab:attacks_properties}
\end{table*}
In addition to the added tokens' realisticity, we consider the added amounts' plausibility for those tokens.
As we show, the transaction's value has little effect on the attack's success, so we only need certainty that the monetary value is realistic and not excessively large.
We can find the histograms for the original and the Sampling Fool adversarial sequences in~\autoref{fig:amounts_age_sf}.
We see that the adversarial sequences meet both of the specified requirements.
For other datasets and attacks, the figures are similar. 

\vspace{-5mm} 
\section{Conclusion}
\vspace{-3mm} 
We consider an essential topic of adversarial attacks on machine-learning models for a highly relevant use case, transactions record data, and present defences against such attacks.

We find out that alone by inserting a couple of transactions in a sequence of transactions, it is possible to fool a model. 
Even small model accuracy drops can lead to significant financial losses to an organization, while for most datasets we observe a severe model quality decrease.
Moreover, attack are the most effective for important borderline decision cases.
However, we can easily repel most classic attacks.
It is straightforward to detect an adversarial sequence or to fine-tune a model to process adversarial sequences correctly.

Attacks stemming from generative data models represent the most promising ones. It is challenging to defend against them, and they show high efficiency. 
During our experiments, a manual assessment of the adversarial sequences show that they are realistic, and thus experts must be careful whenever deploying deep-learning models for processing transaction records data.
Nevertheless, we still can detect adversarial sequences in most realistic scenarios, when an attack appends new transactions only to the end of a sequence.

We expect, that our result shades a light on how attacks and defences should look like for models based on transactions record data. They can be extended to more general event sequences data.

\bibliographystyle{ACM-Reference-Format}
\bibliography{sample-base}
\clearpage
\appendix

\vspace{-5mm} 
\section{Research Methods}
\vspace{-3mm} 
\subsection{Implementation details}

Here, we give an overview of the implementation details. 
We base our implementation on the AllenNlp library.
\begin{enumerate}
    \item \textbf{Reading data}. 
    To read the original transactions record data, and to transform it into the so-called \textit{allennlp.data.fields} format, which we require to manipulate our data, we use the \textit{Transactions Reader}, based on the \textit{DatasetReader} from the AllenNlp library, with a pre-trained discretizer to transform monetary values and further work with these amounts in a discrete space. 
    The discretizer is the \textit{KBinsDiscretizer} with hyperparameters, n bins=$100$, strategy=quantile, encode=ordinal.
    
    \item \textbf{Classifiers}. 
    In this work, we use a bidirectional GRU, an LSTM with $1$ layer, a dropout with $0.1$, a hidden size of $256$ or a CNN with $ 64$ filters and a ReLu activation layer. 
    We use cross-entropy as the loss function. 
    We train all models with a batch size $1024$ for $50$ epochs with an early stop if the loss does not decrease during the validation phase within $3$ epochs. 
    Further, we use the Adam optimizer with a step size $0.001$.
    
     \item \textbf{Attacks}. 
     We conduct most of the experiments with attacks using black-box settings. 
     For our experiments,~\autoref{tab:general_results} is the result of an RNN with a GRU block as an attacked model and an RNN with a GRU block as a substitute model that an attacker uses to assess the so-called \textit{adversality} of the generated sequences. 
     We generate \textbf{1000} adversarial sequences from the test set and average evaluated model performance metrics over them.
     
     All hyperparameters and implementation details for our attacks are in \autoref{sec:attack_methods}.
     We list them as follows:
     \begin{itemize}
         
         \item \textbf{Sampling Fool (SF)}. 
         The algorithm hyperparameters of the Sampling Fool attack include temperature and the number of samples. The temperature regulates how close the distribution of tokens is to be uniform. 
         At high temperatures, attacks are more diverse but less effective. 
         We set the temperature parameter to $2.0$ and the number of samples to $200$.
         
         We use the Masked Language Model (MLM)~\cite{devlin2018bert}, as our language model for the Sampling Fool attack. 
         The MLM accidentally masks some tokens from the original sequence during training and tries to predict which tokens were instead of the masked ones. 
         In our task, we input the model not only of the embeddings of transactions but also the embeddings of the amounts of these transactions, which are concatenated and fed to the encoder. 
         We train the model by maximizing the probability of the sequences.
         $
\max _{\theta} \log p_{\theta}(\mathbf{x})=\sum_{t=1}^{T} \log p_{\theta}\left(x_{t} \mid \mathbf{x}_{<t}\right)
$.

\begin{table*}[t]
    \centering
\begin{tabular}{cccccc}
\hline
Attack & NAD $\uparrow$ & WER $\downarrow$ & AA $\downarrow$  & PD $\uparrow$ \\ \\
\hline
\multicolumn{5}{c}{Age 1} \\
\hline
SF & 0.08 & 12.79 & 0.48 &  0.08 \\
Concat SF & 0.46 & 2 & 0.09 & 0.26\\
FGSM &  0.44 & 1.73& 0.51 &  0.06\\
Concat FGSM & 0.24 & 4 & 0.02 & 0.24  \\
\hline
\multicolumn{5}{c}{Client Leaving} \\
\hline

SF & 0.17 &  14.71 &0.67 & 0.01 \\
Concat SF & 0.18 & 2 & 0.63 & 0.01 \\
FGSM & 0.29 &  8.67 & 0.50 & 0.08\\
Concat FGSM & 0.11 & 4& 0.53 & 0.09 \\
\hline
\multicolumn{5}{c}{Scoring} \\
\hline
SF & 0.12 & 6.16& 0.83 & -0.05 &\\
Concat SF & 0.05 & 6 & 0.67 & 0.07 &\\
FGSM & 0.23 & 8.72 & 0.67 & 0.08 &\\
Concat FGSM & 0.08 & 6 & 0.50 & 0.19 &\\
\hline
\end{tabular}
    \caption{Attacking of model with CNN architecture using substitute model with GRU architecture}
    \label{tab:cnn_vs_gru}
\end{table*}

\begin{figure}[H]
    \begin{center}
    \centerline{\includegraphics[width=0.8\columnwidth]{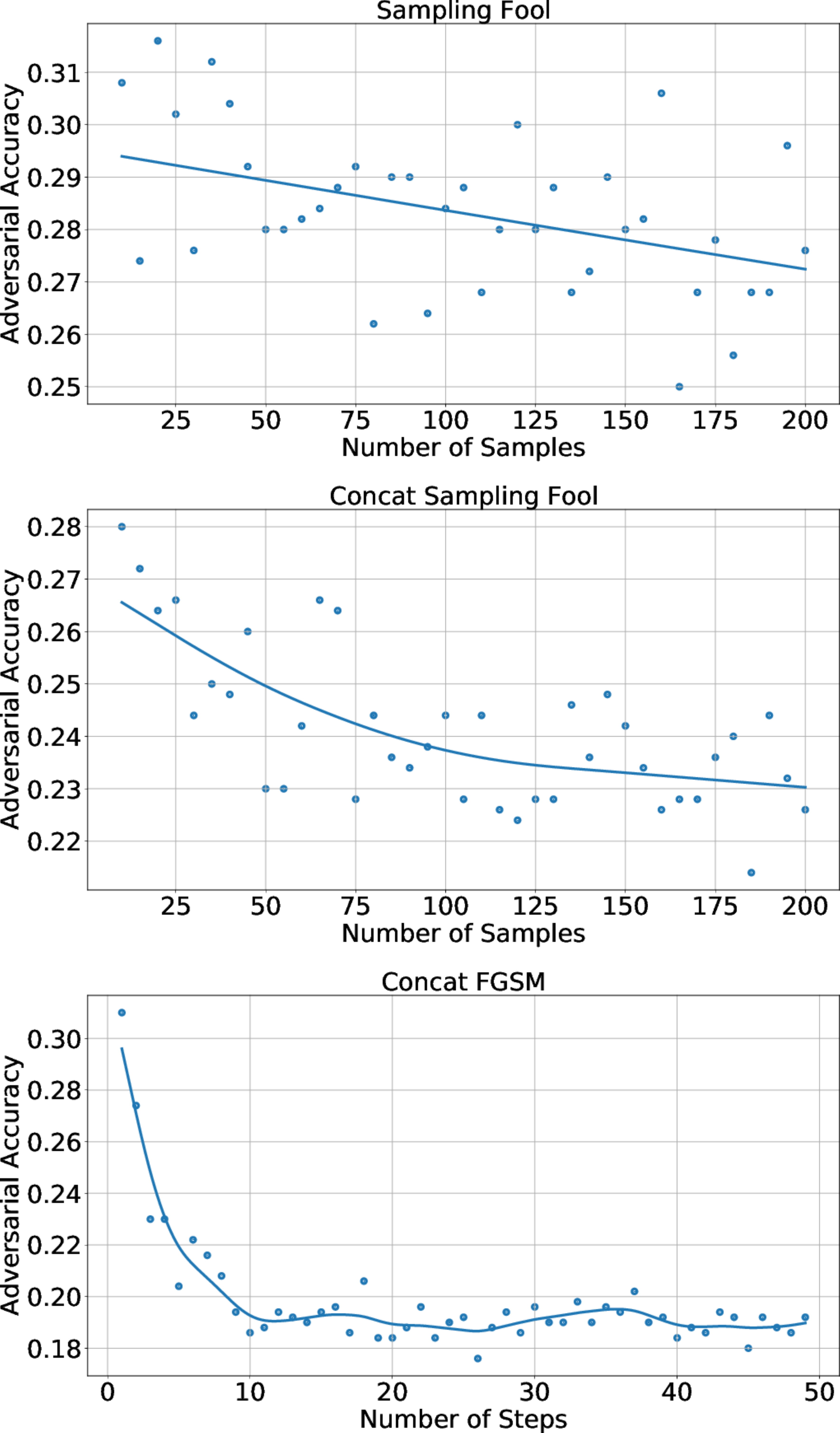}}
    \caption{Dependence of the Adversarial Accuracy Metric on the number of samples for Sampling Fool and Concat Sampling Fool, and a number of iterations in the case of Concat FGSM. For this experiment, we used Age 2 dataset.}
    \label{fig:hyperparameters_num_iterations}
    \end{center}
\end{figure}

We use these models to sample the most plausible sequences from transactions with a constraint on the input amounts:
    \item \textbf{Concat SF, Sequential Concat SF}. 
    The attacks have the same hyperparameters as SF with an additional parameter for the number of tokens, added in the attack. We set this parameter to $2$.
     \item \textbf{FGSM}. 
     We have the hyperparameters, $\epsilon$ for the step size towards the antigradient, and $n$ as the algorithm's number of steps. 
     We set $\epsilon$ to $1.0$ and number of steps to $30$.
     \item \textbf{Concat FGSM}. 
     The attack has the same hyperparameters as FGSM with an additional parameter for the number of tokens added in the attack. 
     We set this parameter to $4$.
     \end{itemize}
     
\end{enumerate}
\vspace{-5mm}
\begin{figure}[H]
    \begin{center}

    \centerline{ \includegraphics[width=0.8\columnwidth]{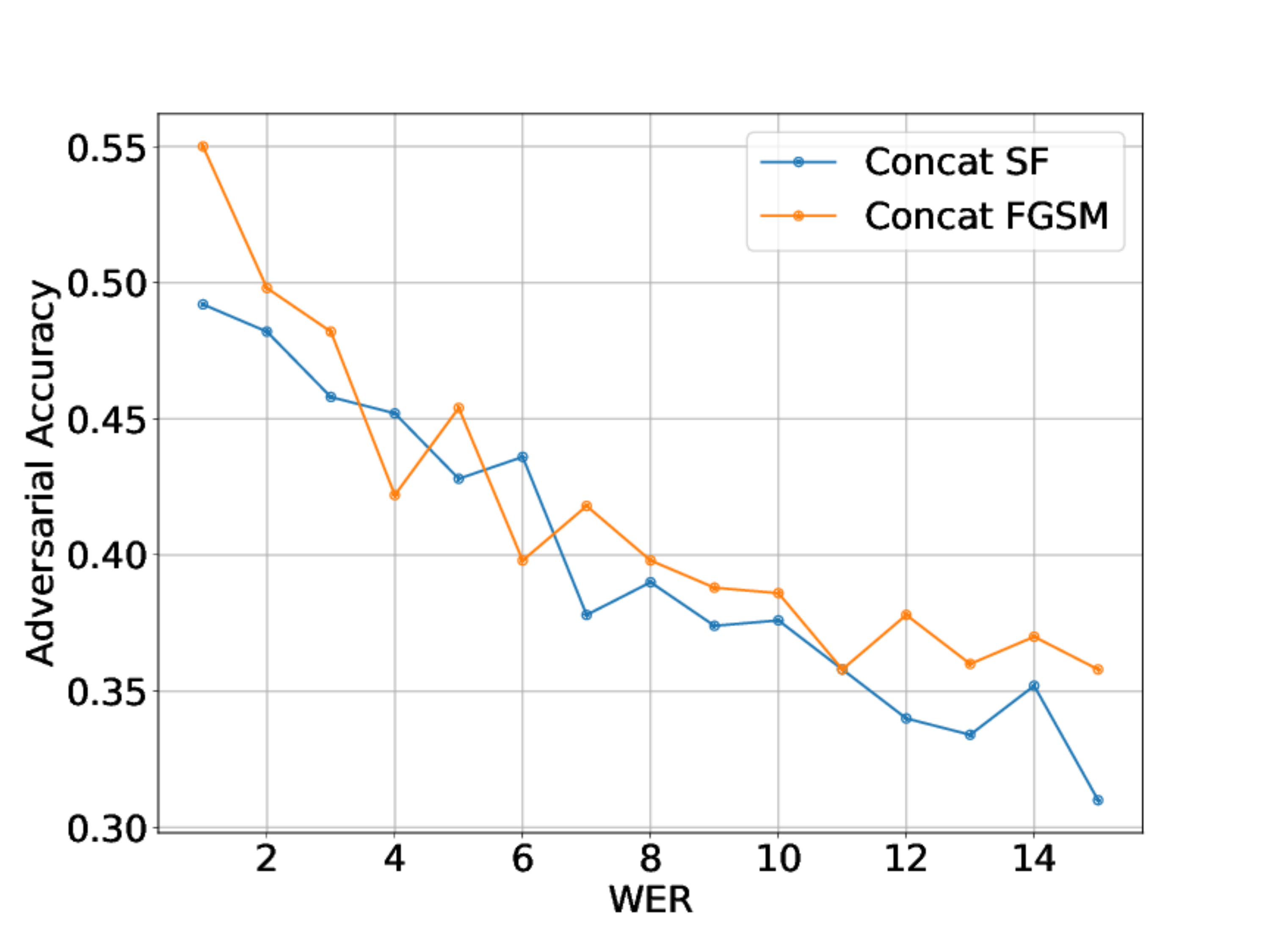}}
    \caption{Dependence of the Adversarial Accuracy Metric on the number of added adversarial tokens (or WER) for Concat SF and Concat FGSM. Client leaving dataset is under consideration, GRU is used as the target and substitute models.}
    \label{fig:dependence_AA_on_ME}
    \end{center}
\end{figure}

\begin{figure}[H]
\begin{center}
  \centerline{\includegraphics[width=\columnwidth]{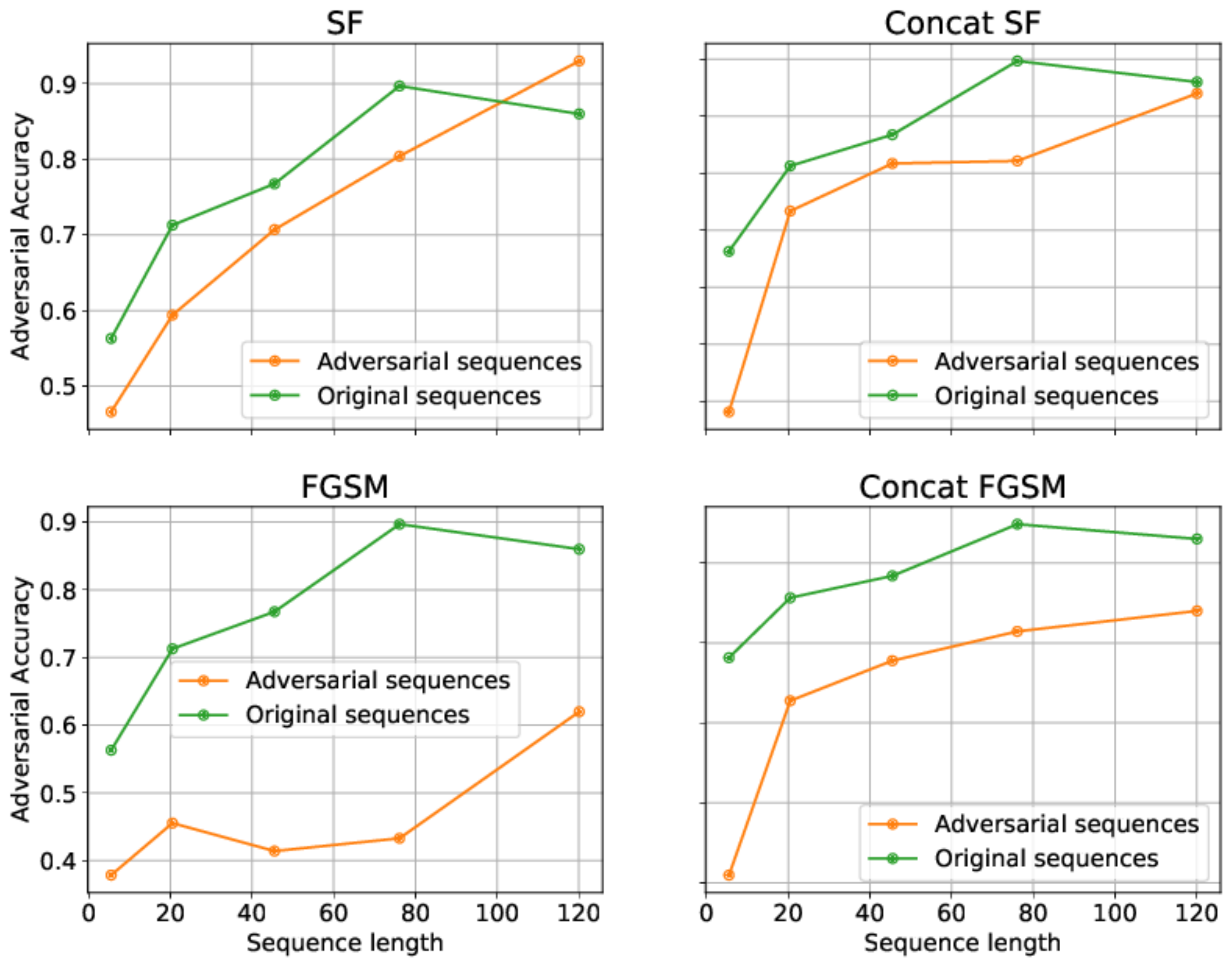}}
    \caption{Dependence of the Adversarial Accuracy Metric on the length of the original sequences for Client leaving dataset. The target and substitute models are GRU. The green line corresponds to the accuracy of the target classifier for the original sequences and the orange line corresponds to the results on adversarial sequences.}
    \label{fig:dependence_aa_on_seq_length}
    \end{center}
\end{figure}

\begin{figure}[H]
\begin{center}
\includegraphics[width=\columnwidth]{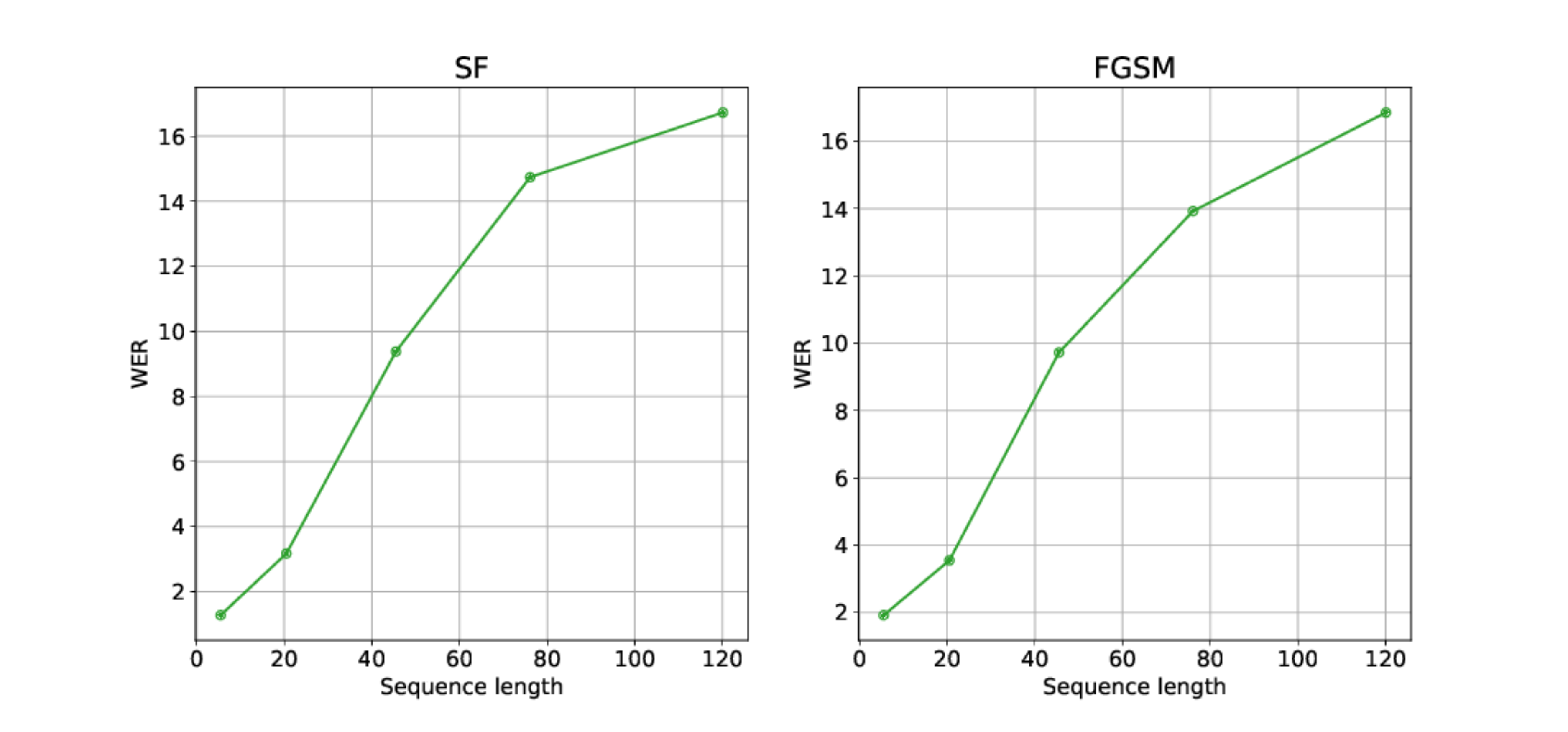}
    \caption{Dependence of WER on length of the original sequences for successful attacks. Client leaving dataset was used. The target and substitute models are GRU.}
    \label{fig:dependence_wer_on_seq_length}
    \end{center}
\end{figure}
\vspace{-5mm}
\subsection{Alternative model architectures}
\vspace{-2mm}
We perform additional experiments for different architectures of the target and substitute models.
The results are in~\autoref{tab:cnn_vs_gru}. 
We can attack the target model consisting of a CNN architecture with a substitute model entailing a GRU architecture, and the quality of some attacks will be still high. We are mainly considering when the target and substitute models are both a GRU architecture. 
However, most of our main claims hold when both models also have a CNN architecture.
\vspace{-5mm}
\subsection{Varying hyperparameters}
\vspace{-2mm}
We select all parameters for the attacks by doing brute force on the grid. 
We examine the efficiency of the attacks by varying the number of sample hyperparameters.
In~\autoref{fig:hyperparameters_num_iterations}, we can observe the results for our considered attacks.
An increase in the number of samples linearly impacts the scale attack time, but we observe that it is sufficient to use about $100$ to reach the attack quality plateau for all attacks.
\vspace{-5mm}
\subsection{Influence of WER on attacks effectiveness}
\vspace{-2mm}
We evaluate the Concat SF and Concat FGSM attacks to understand how the number of appended adversarial tokens at the end affects the attacks' quality. 
In~\autoref{fig:dependence_AA_on_ME}, we depict the result of the dependence of Adversarial Accuracy on WER. 
From this experiment, we can see that we increase the attack's effectiveness by adding more tokens at the end of the sequence. 
\vspace{-5mm}
\subsection{Connection between attacks metrics and sequence lengths}
\vspace{-2mm}
We are also interested in the connection between the attacks' metrics and the length of the initial transaction sequences. 
In~\autoref{fig:dependence_aa_on_seq_length}, we assess the dependence of Adversarial Accuracy on sequence length. 
We divide all transaction records sequences into five groups according to their length. 
Each point corresponds to the center of one of five length intervals. 
Then, on each interval, we record the number of attacks where the initial label does not change. 
From our observations, we conclude that long sequences are more robust to the attacks, while short adversarial sequences are more vulnerable to the attacks because of the absence of long-term patterns in them. 
Also, we want to understand the relationship between the sequence lengths and the number of adversarial tokens required in the sequence to lead to misclassification. 
Our results are in~\autoref{fig:dependence_wer_on_seq_length}. 
We must add more adversarial tokens to the long sequences, so the target classifier generates a wrong prediction compared to the short sequences.


\end{multicols}
\end{document}